\pdfoutput=1

\documentclass[11pt]{article}

\usepackage[]{ACL2023}

\usepackage{times}
\usepackage{latexsym}
 
\usepackage[T1]{fontenc}

\usepackage[utf8]{inputenc}

\usepackage{microtype}

\usepackage{inconsolata}

\usepackage{times}
\usepackage{latexsym}
\usepackage{algorithm,algpseudocode}
\usepackage{subcaption}
\usepackage{hhline}
\usepackage{booktabs}
\usepackage{multirow}
\usepackage{amsmath}
\usepackage{graphicx}
\usepackage{makecell}
\newcommand{\figref}[1]{Figure \ref{#1}}

\newcommand{\tabref}[1]{Table \ref{#1}}
\newcommand{\secref}[1]{Section \ref{#1}}

\usepackage{color}

\title{In-sample Curriculum Learning by Sequence Completion for Natural Language Generation}

\author{Qi Jia$^1$, Yizhu Liu$^2$, Haifeng Tang$^3$, Kenny Q. Zhu$^4$\thanks{\hspace{2mm}The corresponding author.}\\
	$^{1,4}$Shanghai Jiao Tong University, Shanghai, China \\
	$^2$Meituan, Shanghai, China \\
	$^3$China Merchants Bank Credit Card Center, Shanghai, China \\
	\texttt{$^1$Jia\_qi@sjtu.edu.cn},
	\texttt{$^2$liuyizhu@meituan.com} \\ 
	\texttt{$^3$thfeng@cmbchina.com},
	\texttt{$^4$kzhu@cs.sjtu.edu.cn}\\
}

\begin{document}
\maketitle
\begin{abstract}
Curriculum learning has shown promising improvements in multiple domains by 
training machine learning models from easy samples to hard ones. 
Previous works which either design rules or train models for scoring the difficulty 
highly rely on task-specific expertise, and cannot generalize. Inspired by the ``easy-to-hard'' 
intuition, we propose to do in-sample curriculum learning for natural language generation tasks. 
Our learning strategy starts training the model to generate the last few words, i.e., 
do sequence completion, and gradually extends to generate the whole output sequence.
Comprehensive experiments show that it generalizes well to different
tasks and achieves significant improvements over strong baselines.
\end{abstract}

\section{Introduction}
\label{sec:intro}

Curriculum learning (CL) proposed by~\citet{bengio2009curriculum} provides performance improvements on a number of machine learning tasks.
It mimics the learning process of humans by training models with samples in a 
more meaningful order, i.e., from the easy ones to the hard ones.
Therefore, ranking training samples by difficulty lies in the core of CL, 
which is also the key challenge when it's applied to natural language generation (NLG) 
tasks.

Previous work on CL for NLG focuses on measuring the difficulty of training samples 
in two ways. One is to resort to human-crafted rules based on various linguistic features 
and human observations~\cite{liu2018curriculum,kocmi2017curriculum}. 
The other uses models either trained from outside data or the same data but
in previous epochs/steps~\cite{zhou2020uncertainty,kumar2019reinforcement,shen2020cdl}.
Either way seeks to produce a numeric score for each training sample relying on domain expertise so that it can
be ranked,  
making it difficult to generalize 
to different tasks. For example, summarization focuses more on generating concise 
outputs while style transfer emphasizes style changes. So the former should pay 
attention to the ratio between the lengths of the output and the input (the more compressed
the more difficult), while the 
latter should focus on differences in style between the input and output (the more
different the more difficult).
Designing a comprehensive or universal scoring function is difficult or 
even impossible under this definition of CL.
\begin{figure*}[th]
	\centering
	\begin{minipage}[t]{0.33\linewidth}
		\centering
		\subfloat[Sample-wise Curriculum Learning]{
			\includegraphics[scale=0.55]{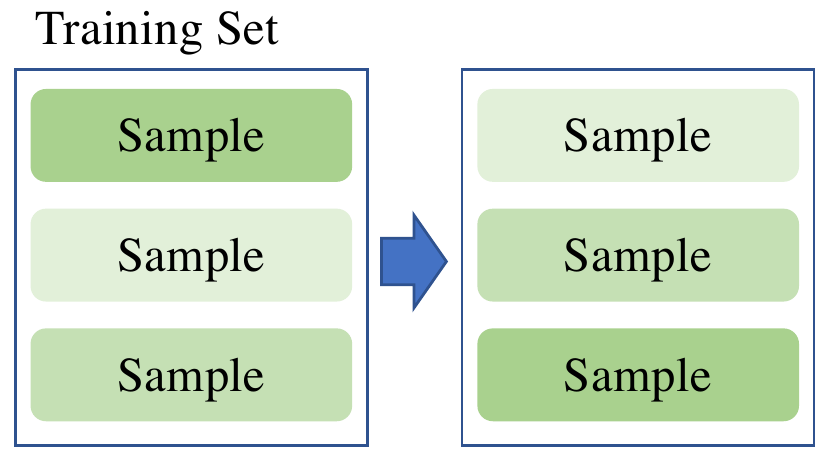}
		}%
	\end{minipage}%
	\begin{minipage}[t]{0.33\linewidth}
		\centering
		\subfloat[Token-wise Curriculum Learning]{
			\includegraphics[scale=0.5]{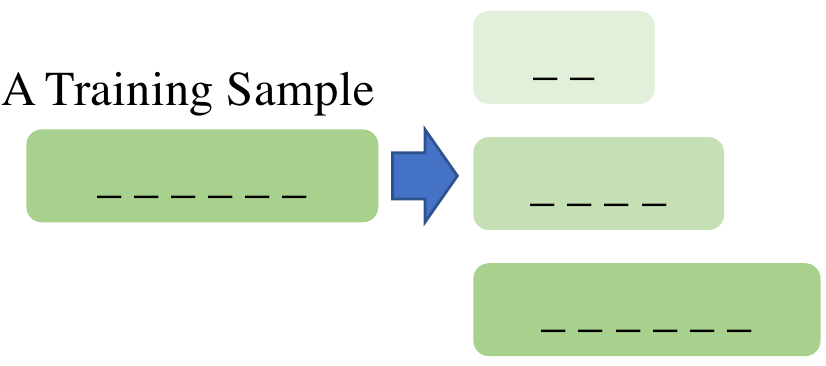}
		}%
	\end{minipage}%
	\begin{minipage}[t]{0.33\linewidth}
		\centering
		\subfloat[In-sample Curriculum Learning]{
			\includegraphics[scale=0.5]{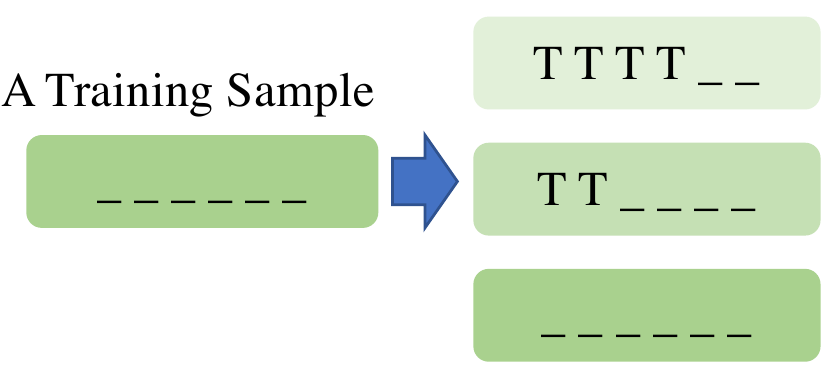}
		}%
	\end{minipage}%
	\centering
	\caption{Illustrations of the main ideas of traditional sample-wise CL, Liang et al.~\shortcite{liang-etal-2021-token-wise}'s TCL and our ICL. 
Green box refers to samples in different difficulty levels. The darker, the harder. ``T'' refers to a known token while ``\_'' refer to a token to be
generated in the output sequence of a sample during training.}
	\label{fig:intro}
\end{figure*}

In this paper, we propose an alternative to sample-wise CL, which we call 
in-sample CL (ICL). 
ICL re-orders the learning sequence within the sample. 
One particular ICL re-ordering strategy which we find effective is to predict the 
last few tokens given a long prefix first from the original output, and then gradually increase the number of tokens
at the end while shortening the prefix, to create an easy-to-hard training
order. Such a curriculum learning strategy focuses more on the difficulty of language generation itself,
leading to a better generalization ability among tasks.

Actually, we are not the first to propose the idea of ICL.
\citet{liang-etal-2021-token-wise} introduced the notion of ``token-wise curriculum learning(TCL)''. 
Illustrations of TCL, ICL and the traditional CL are shown in \figref{fig:intro}.
Their work considers generating the first few tokens
in the output sequence to be easier than generating a longer sequence in the output. 
Based on this idea, they proposed a ``hard'' version of TCL that creates training samples of increasing output length by cutting the sentences short. 
In this way, TCL is similar to data augmentation with incomplete and even ``incorrect'' samples, while our ICL considers each training sample in full
length. A ``soft'' version of TCL that places decaying weights on the end tokens instead of cutting short is introduced as a mitigation to avoid incomplete samples,
which was proved to uniformly outperform the ``hard'' version.

To validate the advantage of ICL, we conduct extensive experiments on a range of natural language generation tasks, including reading comprehension, dialogue summarization, style transfer, question generation and news summarization, with different backbone models, such as BART, UniLM and GPT-2. The results show the favorable performance of ICL over the strong baselines.
In a word, our contributions are:
\begin{itemize}
	\item We propose an improved in-sample curriculum learning strategy for text generation by doing sequence completion (\secref{sec:icl}).
	\item We propose a novel ICL learning algorithm (\secref{sec:iclalgorithm}). 
Together with our sequence completion ICL curriculum, it achieves significant improvements over 
the strong baselines on different NLG tasks, demonstrating strong generalization ability
(\secref{sec:experiment}).
	\item Our approach can be combined with traditional CL for further performance gains 
(Section~\ref{sec:tracl}). 
\end{itemize}

\section{Approach}
\label{sec:approach}
We present an ICL strategy in the context of the vanilla sequence-to-sequence (Seq2Seq) 
training objective with a detailed learning algorithm.

\subsection{ICL by Sequence Completion}
\label{sec:icl}

Today, NLG tasks are generally solved by Seq2Seq models, especially the pre-trained language models. 
Vanilla Seq2Seq models are trained to predict the output $Y=\{ y_1, ..., y_n\}$ 
given the input $X$ by minimizing the negative log-likelihood:
\begin{equation}
L_{orig} = -\frac{1}{n}\sum_{t=1}^{n}\log P(y_t|y_{<t}, X)
\end{equation}
Traditional CL manipulates the selection of training pair $(X, Y)$ from easier pairs to harder ones for different tasks with this vanilla loss function.

In contrast, ICL digs into the output sequence itself and exploits the difficulty of 
language generation within each training sample.
We segment $Y$ into two sub-sequences by a cutting point $c$, where $1\leq c\leq n$. The sub-sequence 
before $c$ is called the \textit{prefix}, and the one after (and including) $c$ is the \textit{target}. 
According to the Shannon Information Theory, the entropy goes down when more related information is given.
Thus, the difficulty of the sequence completion task that generates the target will decrease when a longer prefix is given. In other words, we can manipulate $c$ to vary the difficulty of samples during 
training.

\begin{figure}[th]
 	\centering
 	\includegraphics[scale=0.45]{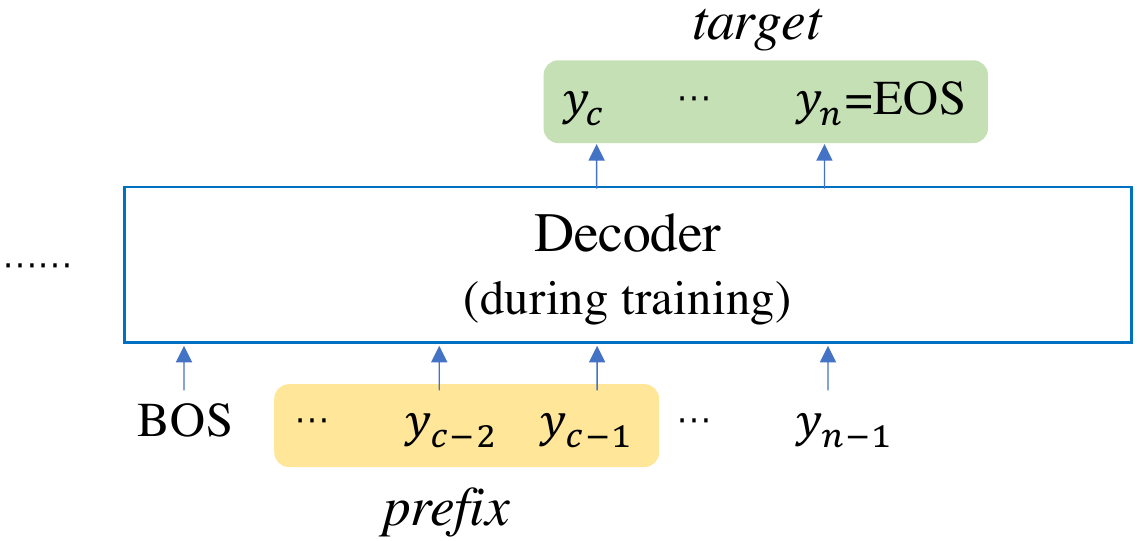}
 	\caption{The decoder of a NLG model. BOS and EOS are special tokens representing the beginning and the end of the output. It's suitable for different model architectures, including encoder-decoder, decoder-only, etc.} 
 	\label{fig:decoder}
 \end{figure}

Based on this intuition, we modify the vanilla loss as:
\begin{equation}
	L_{icl} = -\frac{1}{n-c+1}\sum_{t=c}^{n}\log P(y_t|y_{<t}, X)
	\label{eq:icl}
\end{equation}
i.e., given $X$ and the prefix as inputs to the encoder and decoder respectively, we only calculate the loss for predicting the target.
At the beginning of the training process, we use a larger $c$ to train the model to predict 
the target with only the last few words.
Then, we gradually decrease $c$, until the prefix reduces to an empty sequence.
In this way, the model grows stronger with more difficult generation objectives and 
learns to generate the whole output in the end. An illustration is in Figure~\ref{fig:decoder}.

\subsection{ICL Algorithm}
\label{sec:iclalgorithm}
Since the output length varies from sample to sample, it's hard to set $c$ as a constant for all samples.
If so, samples with short outputs will be neglected when $c$ is large at the beginning, and the model will eventually bias to training samples with long outputs as they are shown more times.
In light of this, we proposed to determine $c$ sample by sample relative to their output lengths.

We define a start point $p_{start}$ and a stride $s$ for controlling $c$, where $0\leq p_{start}, s \leq 1$.
The training process starts with: 
\begin{equation}
	c = n\times p_{start}
	\label{eq:cnp}
\end{equation}
After each epoch or a number of updating steps, we validate the model on the validation set. If the performance on the validation set no longer increases, we introduce a more difficult generation task 
by removing $s$ from $p_{prev}$:
\begin{equation*}
	p_{new} = 
	\begin{cases}
	p_{prev}-s, & \text{if $p_{prev}>s$} \\
	0, & \text{else}
	\end{cases} \\
\end{equation*}
and update $c$ by Equation~\ref{eq:cnp}. The training process terminates until there are no improvements on the validation set with $c$ equaling 0.
More details are included in Algorithm~\ref{alg:picl}. 

\begin{algorithm}[th]
	\caption{The ICL training algorithm.}
	\label{alg:picl}
	\small
	\textbf{Input}: the model to be fine-tuned $M_{in}$, the training set $D_t$, the validation set $D_v$\\
	\textbf{Parameter}: a start point $p_{start}$, a stride $s$\\
	\textbf{Output}: the final model $M_{out}$
	\begin{algorithmic}[1] 
		\Procedure{ICL}{$M_{in}, D_t, D_v, p_{start}, s$}	
		\State $p = p_{start}$ 
		\State $M_{out}=M_{in}$ 
			
		\For{training epoch $e=1,...$}
			\State \Comment{Training process}
			\For{training steps in an epoch}
				\State Randomly sample a batch $B$ from $D_t$
				\For{Each sample $(X, Y)$ in $B$}
					\State $c = n\times p$
					\State Calculate $L_{icl}$ by Eq.~\ref{eq:icl}
				\EndFor
				\State Update $M_{in}$ based on $\frac{1}{|B|}\sum_{|B|}L_{icl}$
			\EndFor
			\State \Comment{Validation process}
			\State Calculate $M_{in}$'s performance on $D_v$.
			\If{$M_{in}$ gets improvements on $D_v$}
				\State $M_{out} = M_{in}$
			\Else
				\State Update $p$ according to Eq.~\ref{eq:cnp}
			\EndIf
		\EndFor	
		\State \textbf{return} $M_{out}$
	
		\EndProcedure
	\end{algorithmic}

\end{algorithm}


\section{Experiment}
\label{sec:experiment}
In this section,
we first present the experimental setups for different tasks. Then, we show the quantitative and qualitative results together with comprehensive analysis and case studies.

\begin{table*}[th]
	\scriptsize
	\centering
	\begin{tabular}{lp{1.1cm}p{0.9cm}rrrcccc}
		\hline
		Task & Dataset & Model & \#Train & \#Val & \#Test & Input & Output & Avg & Std\\
		\hline
		Reading Comprehension & DREAM &BART& 6,116 & 2,040 & 2,041 & ``Q:''+ question + dialogue & answer & 5.59 & 2.61\\
		Dialogue Summarization & SAMSum &BART& 14,732 & 818 & 819 & dialogue & summary  & 24.99 & 13.06\\
		\multirow{2}{*}{Style Transfer} & \multirow{2}{*}{Shakespeare} &{STRAP}&  \multirow{2}{*}{36,790} &  \multirow{2}{*}{2,436} &  \multirow{2}{*}{2,924} & {original}  & {modern}  &  \multirow{2}{*}{11.63} & \multirow{2}{*}{8.19} \\
		&&(GPT-2) &  & && /modern& /original & & \\
		Question Generation & SQuAD1.1 &UniLM& 75,722 & 10,570 & 11,877 & passage + [SEP] + answer & question & 13.09 & 4.27 \\
		News Summarization & CNNDM &BART& 287,227& 13,368& 11,490 & document & summary & 70.97 & 29.59\\ 
		\hline
	\end{tabular}
	\caption{A summary of tasks and datasets. \#Train, \#Val and \#Test refers to the number of samples in the corresponding dataset. Avg and Std are the statistics for the number of output tokens. ``+'' is the concatenation operation.}
	\label{tab:taskdata}
\end{table*}
\subsection{Experimental Setups}
\label{sec:implementation}
We did experiments on five commonly-researched natural language generation tasks as follows:

\textbf{Reading comprehension} is the task that answering questions about a piece of text. We use the DREAM dataset~\cite{sun2019dream} where questions are about corresponding dialogues and the answer is a complete sentence in natural language. We neglect the negative choices in the original dataset and formulate it as a NLG task. We adopt the pre-trained language model BART~\footnote{\url{https://huggingface.co/facebook/bart-large}}~\cite{lewis2020bart} as the baseline.
The generated answers are evaluated by BLEU scores
~\cite{papineni2002bleu} widely used for QA systems, together with Meteor and Rouge-L F1~\cite{fabbri2021summeval}.
We evaluate the model after each training epoch and the early-stop patience will be added 1 if there is no improvement 
in the perplexity on the validation set. The training process terminates when the early-stop patience equals or is larger than 3.  
During the inference, the minimum and maximum output length are set to 5 and 100, with no\_repeat\_ngram\_size=3, length\_penalty=1.0 and num\_beams=4.

\textbf{Dialogue summarization} is to generate a concise summary covering the salient information in the input dialogue. The preceding model BART has shown to be a strong baseline for this task.
We experiment with  
SAMSum dataset
~\cite{gliwa2019samsum} for daily-chat dialogues. 
The generated summaries are evaluated by comparing with the reference through evaluation metrics, including Rouge-1/2/L F1 scores 
~\cite{lin2004rouge}, Meteor~\cite{banerjee2005meteor} and BertScore F1. 
The parameters are the same as reading comprehension, except that the early-stop is activated if there is no improvement according to the Rouge-2 F1 score.

\textbf{Style transfer} preserves the semantic meaning of a given sentence while modifying its style, such as positive to negative, formal to informal, etc.
We adopt the Shakespeare author imitation dataset~\cite{xu2012paraphrasing}, containing William Shakespeare's original plays and corresponding modernized versions. Krishna et al.~\shortcite{krishna2020reformulating} proposed to do unsupervised style transfer by training paraphrase models based on the GPT-2 language model~\cite{radford2019language}. We re-implemented their approach STRAP. 
Evaluation metrics include
transfer accuracy(ACC), semantic similarity(SIM), Fluency(FL) and two aggregation metrics, i.e., geometric averaging(GM) and their proposed $J(\cdot)$ metric~\footnote{GM calculate the geometric mean of the corpus-level ACC, SIM and FL. $J(\cdot)$ first calculates the multiplication of sample-level ACC, SIM and FL, then get the average score across the test corpus.}. 
In the training stage, we evaluate the model after updating every 500 steps. The perplexity on the validation set is used to activate the early-stop which equals 3. The inference is done as default.
 
\textbf{Question generation}~\cite{zhou2017neural} aims at generating a question given an input document and its corresponding answer span. SQuAD 1.1~\cite{rajpurkar2016squad} is generally used for evaluation. We adopt the data split as in \cite{du2017learning} and fine-tune the pre-trained UniLM~\cite{dong2019unified} as the strong baseline. 
Generated questions are evaluated by metrics including BLEU-1/2/3/4, Meteor and Rouge-L with the provided scripts. The model is evaluated every 1000 steps and the early-stop equaling 5 is associated with the perplexity on the validation set. Other parameters are unchanged following the official guideline.

\textbf{News summarization} differs from dialogue summarization where the input is a document instead of a dialogue. We adopt the same strong baseline BART and evaluation metrics as dialogue summarization. Experiments are done with CNNDM dataset~\cite{HermannKGEKSB15} consisting of news articles and multi-sentence summaries. The model is evaluated every 3000 steps and the early-stop equaling 3 is associated with the Rouge-2 on the validation set. During the inference, the minimum and maximum output length is set to 45 and 140 respectively, with no\_repeat\_ngram\_size=3, length\_penalty=2.0 and num\_beams=4 \footnote{Inference parameters are borrowed from \url{https://github.com/pytorch/fairseq/blob/main/examples/bart/summarize.py}.}.

A summary of these tasks is in Table~\ref{tab:taskdata} and the specific packages we adopted are in the Appendix. 
For fair comparisons, we re-implement these baselines on our machine. Then, we further arm them with different in-sample curriculum settings without changing corresponding hyper-parameters. Specifically, we distinguish \citet{liang-etal-2021-token-wise}'s work and our method in detail from two aspects, including the curriculum criterion denoted by SG or SC and the training algorithm denoted by TCL or ICL~\footnote{In the rest sections, we use TCL and ICL to refer to the corresponding training algorithms specifically.}, which results in the following 4 combinations:
\begin{itemize}
	\item \textbf{TCL-SG}: the token-wise curriculum learning algorithm(TCL) with sub-sequence generation(SG) criterion proposed by \citet{liang-etal-2021-token-wise} with their best soft setting. The hyper-parameters are set as $\gamma_0=0.7$ and $\alpha_0=25$ following the original paper.
	\item \textbf{TCL-SC}: we modified the TCL-SG by incorporating our sequence completion(SC) criterion in \secref{sec:approach} with the hard setting~\footnote{The soft setting will hurt our ordering criterion according to preliminary studies in Appendix.} where $\lambda_0=0.1$ following the original paper.
	\item \textbf{ICL-SG}: we implemented the SG criterion by using our ICL algorithm in \secref{sec:approach} which calculating the loss with $1\leq t \leq c$ in \eqref{eq:icl}.
	\item \textbf{ICL-SC}: our final approach. Both TCL-SC and ICL-SG are ablations for it. The settings of newly introduced $p_{start}$ and $s$ are specified and discussed in Section~\ref{sec:params}.
\end{itemize}

All of the approaches are trained with the same max training epochs with the early-stop for preventing from over-fitting.
The experiments are done on a single RTX 3090 with 24G GPU memory. The results are averaged over three runs. We open-source all of codes and results at \url{https://github.com/JiaQiSJTU/InsampleCurriculumLearning}. 

\subsection{Automatic Evaluations on Different Tasks}
\label{sec:taskperformances}

The performances on different NLG tasks are shown in Table~\ref{tab:end2end}. 
These tasks not only focus on solving different problems, but also has a various amount of training data as well
as output lengths according to
Table~\ref{tab:taskdata}.
Besides, the basic models are also different, including BART, GPT-2 and UniLM. 
Our approach \textbf{ICL-SC achieves significant improvements over the strong baselines} among different tasks on most evaluation metrics, which shows that our method not only works well, but also has strong generalization abilities.
It should be noted that GM and J are two comprehensive evaluation metrics for style transfer, with our approach topping the ranks with significant improvements.

\begin{table}[t]
	\small
	\centering
	\begin{subtable}{\linewidth}
		\scriptsize
		\centering
		\begin{tabular}{lcccccc}
			\hline
			{Method} & {B1} & {B2} & {B3} & {B4} & {Met} & {RL}\\
			\hline
			w/o CL &  32.03 & 16.01 & 8.77 & \textbf{4.80} & 19.84 & 38.89\\
			TCL-SG &  32.35 & 16.38 & 8.86 & 4.69 & 19.95 & 39.27 \\
			\hline
			TCL-SC & 33.44 & 16.90 & 8.93 & 4.66 & 20.45 & 40.55\\
			ICL-SG & 32.80 & 16.32 & 8.88 & 4.75 & 19.96 & 39.72 \\
			ICL-SC &  \underline{\textbf{33.99}} & \underline{\textbf{17.43}} & {\textbf{9.18 }}& 4.64 & \underline{\textbf{20.60}}& \underline{\textbf{40.78}}\\
			\hline
		\end{tabular}
		\caption{Reading Comprehension}
		\label{tab:end2endrc}
	\end{subtable}
	\\[5pt]
	\begin{subtable}{\linewidth}
		\scriptsize
		\centering
		\begin{tabular}{lccccc}
			\hline
			{Method} & {R1} & {R2} & {RL} & {Met} & {BertS} \\
			\hline
			w/o CL & 51.88 & 27.30 & 42.77 & 24.75 & 71.38 \\
			TCL-SG & 52.43 & 27.65 & 43.56 & 25.17 & 71.86 \\
			\hline
			TCL-SC & 52.69 & \textbf{28.28 }& 43.89 & 25.08 & 71.95 \\
			ICL-SG & 52.95 & 28.07 & \textbf{43.91} & 25.67 & 72.01 \\
			ICL-SC & \underline{\textbf{53.07}} & \underline{28.23} & {43.83} & \underline{\textbf{26.12}}& {\textbf{72.17}} \\
			
			\hline
		\end{tabular}
		\caption{Dialogue Summarization}
		\label{tab:end2endds}
	\end{subtable}
	\\[5pt]
	\begin{subtable}{\linewidth}
		\scriptsize
		\centering
		\begin{tabular}{lccccc}
			
			\hline
			{Method} &  {ACC} & {SIM} & {FL} & {GM} & {J}\\
			\hline
			w/o CL & 70.49 & 55.70 & 85.98 & 69.63 &33.72 \\
			TCL-SG & \textbf{76.09} & 53.79 & 82.97 & 69.76 & 34.02 \\
			\hline
			TCL-SC& 73.27 & 54.84 & 85.49 & 70.03 & 34.56\\
			ICL-SG & 74.60 & 55.75 & 84.89 & \textbf{70.68} & 35.64\\
			ICL-SC & 73.72 & \textbf{55.91} & \textbf{86.30} & \underline{{70.60}} & \underline{\textbf{35.81}} \\
			\hline
		\end{tabular}
		\caption{Style Transfer.}
		\label{tab:end2endst}
	\end{subtable}
	\\[5pt]
	\begin{subtable}{\linewidth}
		\scriptsize
		\centering
		\begin{tabular}{lcccccc}
			\hline
			{Method} & {B1} & {B2} & {B3} & {B4} & {Met} & {RL}\\
			\hline
			w/o CL & {50.36} & 35.81 & 27.46 & 21.62 & 24.56 & 50.88 \\
			TCL-SG &{50.47} & 35.96 & 27.57 & 21.69 & 24.66 & 50.76\\
			\hline
			TCL-SC & 50.48 & 36.04 & 27.67 & 21.79 & 24.70 & 51.17\\
			ICL-SG & 50.89 & 36.28 & 27.83 & 21.92 & 24.82 & 51.16\\
			ICL-SC &  \underline{\textbf{51.02}} & \underline{\textbf{36.39}} & {\textbf{27.90}} & \textbf{21.96} & \textbf{24.90} & \underline{\textbf{51.29}} \\
			\hline
		\end{tabular}
		\caption{Question Generation}
		\label{tab:end2endqg}
	\end{subtable}
	\\[5pt]
	\begin{subtable}{\linewidth}
		\scriptsize
		\centering
		\begin{tabular}{lccccc}
			\hline
			{Method} & {R1} & {R2} & {RL} & {Met} & {BertS}\\
			\hline
			w/o CL &  43.07 & 20.01 & 35.94 & \textbf{21.44} & 63.72 \\
			TCL-SG & 43.03 & 20.19 & 36.22 & 19.58 & 63.84 \\
			\hline
			TCL-SC & 43.63 & 20.69 & 36.70 & 19.84 & 64.18 \\
			ICL-SG & \textbf{43.76} & \textbf{20.81} & \textbf{36.88} & 19.69 & \textbf{64.31}\\
			ICL-SC & \underline{{43.60}} & \underline{{20.66}} & \underline{{36.73}} & 19.64 & \underline{{64.20}}\\
			\hline
		\end{tabular}
		\caption{News Summarization}
		\label{tab:end2endns}
	\end{subtable}
	\caption{Results on different NLG tasks. w/o CL and TCL-SG are two previous strong baselines. 
Both TCL-SC and ICL-SG are variations of our final approach ICL-SC. Scores underlined of ICL-SC 
are statistically significantly better than both baselines in the first two lines with $p<0.05$ 
according to the t-test. }	
	\label{tab:end2end}
\end{table}
To disentangle factors of learning curriculum and training algorithms, we conduct variations of ICL-SC for detailed comparisons to TCL-SG. More observations are as follows. 

\textbf{$\ast$-SC outperforms $\ast$-SG} for both training algorithms, showing that our proposed sequence completion curriculum is a more effective way of doing curriculum learning within a single sample. 
The only exception is that ICL-SG performs better than ICL-SC for news summarization in \tabref{tab:end2endns}. The reason is that multi-sentence summaries in CNNDM are more extractive and cover different salient information in each sentence. Human agreement on salient information is relatively low as shown in \tabref{tab:humaneval}. Consequently, the prefix of a summary can also be a reasonable and more concise 
reference summary with one or more complete sentences. The nature of $\ast$-SG happens to take advantage
of this property.

\textbf{ICL-$\ast$ is better than TCL-$\ast$} with better performance and less computational costs. For TCL training algorithm adopted in \citet{liang-etal-2021-token-wise}, it separates the whole training process into curriculum and ordinary training. The curriculum length is an important hyper-parameter that is required to be estimated by finishing the training of the baseline model and computing the number of steps it takes to reach approximately 70\% of final scores. It intensively aggravates the computational costs. Besides, this estimation rule can not generalize well to different tasks (More in Appendix). We choose to set curriculum steps to 2 or 3 epochs, approximately to the same amount of samples with different difficulty levels in ICL-SC. Taking dialogue summarization as an example, TCL-SG takes around 15.67 epochs (6 for the curriculum step estimation, 3 for curriculum and 6.67 for ordinary training) while our ICL-SC takes only 11.67 epochs to get the final results (More in Appendix). In a word, our ICL-$\ast$ do the curriculum and ordinary training in a unified manner, requiring less computational costs in total.
Moreover, ICL-$\ast$ moves to the next difficulty level after the model has fully been trained on that judging by the performance on the validation set, which is more similar to the education process in real life and leads to better results.

\subsection{Human Evaluations}

To further prove the improvement of our approach, we asked three proficient English speakers from Asia for human evaluation. 100 samples from the test set of each task are randomly selected, ignoring the ones with totally same generations among three models, including the vanilla model, TCL-SG and ICL-SC. The original input, reference output and three generations are shown to annotators together, while the order of the three generations is unknown and different among samples. 3-point Likert Scale is adopted for scoring each generation~\cite{gliwa2019samsum}, where [5, 3, 1] represent 
excellent, moderate and disappointing results 
respectively. The average scores and annotator agreements 
are in 
Table~\ref{tab:humaneval}.
\begin{table}[h]
	\scriptsize
	\centering
	\begin{tabular}{l|ccc|c}
		\hline
		{Tasks} & {w/o CL} & {TCL-SG} & {ICL-SC} & {Agree}  \\
		\hline
		Reading Comprehension  &3.42 & 3.39 &3.94 &0.64 \\
		Dialog Summarization &3.01 &3.51 &3.6 &0.41 \\
		Style Transfer &2.85 &2.67 & 3.02&0.43 \\
		Question Generation &3.77 & 3.81 &3.93 &0.40 \\
		News Summarization & 3.13 &3.04 &3.43 &0.23 \\
		\hline
	\end{tabular}
	\caption{Human evaluations. The agreement (Agree) is calculated by Fleiss Kappa.}
	\label{tab:humaneval}
\end{table}

The Fleiss Kappa on the first four tasks indicates moderate agreements. It shows the promising improvement of ICL-SC over the vanilla model and TCL-SG, which is consistent with the conclusion based on automatic metrics. The poor agreement on news summarization reflects the 
diverse concerns of summarization from different 
annotators. 

The drop of TCL-SG over the baseline on style transfer is apparent. Although TCL-SG achieves significant improvements in accuracy, the generated contents with less semantic similarities and poor fluency are not preferred by annotators. Examples will be discussed in \secref{sec:casestudies}.

\begin{table}[h!]
	\scriptsize
	\centering
	\begin{subtable}{\linewidth}
		\scriptsize
		\centering
		\begin{tabular}{p{0.9cm}p{5.9cm}}
			\toprule[1pt]
			{Dialogue} & \makecell[l]{
				M: ... So \textbf{health is more valuable than anything else}. No\\\quad  matter how much money we have, ...\\
				W: ... \textbf{honors can never} \textbf{equal good health either.}\\
				M: ... we should try our best to keep us as healthy as possible.\\
			} \\
			\hline
			Question & Which of the following did the man think the most important?\\
			\hline
			Reference & Good health.\\
			\hline
			w/o CL & \textbf{\textit{Honors}}. \\
			\hline
			TCL-SG & \textbf{\textit{Honors}} and health. \\
			\hline
			ICL-SC & Health.\\
			\bottomrule[1pt]
		\end{tabular}
		\caption{Reading comprehension}
		\label{tab:caserc}
	\end{subtable}
	\\[3pt]
	\begin{subtable}{\linewidth}
		\scriptsize
		\centering
		\begin{tabular}{p{0.9cm}p{5.9cm}}
			\toprule[1pt]
			{Dialogue} & \makecell[l]{Mike: <file\_photo> \textbf{woke up} like this :/\\
				Emma: omg what is this??? \textbf{allergy}?\\
				Mike: no idea... probably... but \textbf{no idea} to what :/} \\
			\hline	
			{Reference} & Mike suspects he might have had an allergic reaction to something.\\
			\hline
			
			w/o CL & Mike \textbf{woke up} like \textit{\textbf{this}}.\\
			\hline
			TCL-SG & Mike has an \textbf{allergic reaction}.\\
			\hline
			ICL-SC & Mike \textbf{woke up} with an \textbf{allergic reaction}.\\
			\bottomrule[1pt]
		\end{tabular}
		\caption{Dialogue summarization}
		\label{tab:caseds}
	\end{subtable}
	\\[3pt]
	\begin{subtable}{\linewidth}
		\scriptsize
		\centering
		\begin{tabular}{p{0.9cm}p{5.9cm}}
			\toprule[1pt]
			{Modern} & \makecell[l]{Excuse me , sir , do you know how to read ?} \\
			\hline	
			{Original} &I pray , sir , can you read ?\\
			\hline	
			w/o CL &I \textbf{pray you} , can you read ?\\
			\hline
			TCL-SG & I \textbf{prithee} , \textit{\textbf{read ?}}\\
			\hline
			ICL-SC & I \textbf{prithee} , \textbf{sir} , can you read ?\\
			\midrule[1pt]
			{Original} & \makecell[l]{My \textbf{dismal} scene I needs must act alone .} \\
			\hline	
			{Modern} & In my \textbf{desperate} situation , I have to act alone .\\
			\hline
			w/o CL &I have to act alone in my \textbf{gloomy} scene .\\
			\hline
			TCL-SG &\textbf{\textit{ It's my own fault , my own fault , that I'm the one who's in a d}}\\
			\hline
			ICL-SC & I have to act alone in my \textbf{gloomy} scene .\\
			\bottomrule[1pt]
		\end{tabular}
		\caption{Style Transfer}
		\label{tab:casest}
	\end{subtable}
	\\[3pt]
	\begin{subtable}{\linewidth}
		\scriptsize
		\centering
		\begin{tabular}{p{0.9cm}p{5.9cm}}
			\toprule[1pt]
			{Document} & {Plymouth is home to Plymouth Argyle F.C., who play in \textbf{the fourth tier of English football league known as Football League Two.} ...}\\
			\hline	
			Answer & Football League Two\\
			\hline
			{Reference} & What level of the football league does Plymouth Argyle F.C. operate in?\\
			\hline
			w/o CL & What is the fourth tier of English football league ?\\
			\hline
			TCL-SG & What is \textit{\textbf{the fourth tier of English football }}?\\
			\hline
			ICL-SC & What is the fourth tier of English football league called ? \\
			\bottomrule[1pt]
		\end{tabular}
		\caption{Question Generation}
		\label{tab:caseqg}
	\end{subtable}
	\\[3pt]
	\begin{subtable}{\linewidth}
		\scriptsize
		\centering
		\begin{tabular}{p{0.9cm}p{5.9cm}}
			\toprule[1pt]
			{Document} & {a man has been arrested ... an imam found dead in his car . || abdul hadi arwani was found slumped in the back seat of his black volkswagen passat on tuesday morning in wembley , north west london . || the 48-year-old syrian national was an outspoken critic of the assad regime and ` actively ' campaigned against extremist , his family have since revealed . || on monday morning scotland yard confirmed that a 46-year-old had been arrested in brent , north west london , on suspicion of conspiracy to murder || ... || he is being questioned ... while officers ... for witnesses || ... || counter-terrorism investigators were drafted ...}\\
			\hline
			{Reference} & abdul hadi arwani was found dead in his car on tuesday in wembley . counter terrorism police were drafted in to lead investigation into death . a 46-year-old man has been arrested on suspicion of conspiracy to murder .\\
			\hline
			w/o CL & abdul  ...  london. The 48 ... revealed. A man ... car.\\
			\hline
			TCL-SG & abdul ... london. the 48... revealed. on monday ... arrested on suspicion of conspiracy to murder. he ... witnesses.\\
			\hline
			ICL-SC & abdul ... back seat of his car in wembley. the syrian national was ... assad regime. a 46-year-old man has been arrested on suspicion of conspiracy to murder. he ... witnesses. \\
			\bottomrule[1pt]
		\end{tabular}
		\caption{News Summarization}
		\label{tab:casens}
	\end{subtable}
	\caption{Case studies. 
		\textbf{Keywords} are in bold. \textbf{\textit{Doubtful generations}} are italic. `||'' marks sentence boundaries. Unnecessary words in the document and identical words among generations are folded with ``...''. }
	\label{tab:cases}
\end{table}

\subsection{Case Studies}
\label{sec:casestudies}

We show some cases in Table~\ref{tab:cases}.

In the first case from reading comprehension, our ICL-SC reasoned correctly while the baseline model raised a wrong answer. TCL-SG also answered incorrectly by merging both keywords. 
Such ability is not suitable for generating a precise answer.
In contrast, ICL-SC successfully incorporated more salient information in a single sentence for dialogue summarization, which performs better than both baselines. The vanilla model did poorly on coreference resolution among dialogue utterances and generated ``this'' without a clear referent. 
ICL-SC also generated a more accurate question in Table~\ref{tab:caseqg} compared with strong baselines, although it's not the same as the reference.

For transferring style from modern to Shakespeare's style, the model generated results are all acceptable while ICL-SC performs slightly better for being more polite. Both TCL-SG and ICL-SC even generated the more professional word ``prithee'' which is widely used in Shakespeare's time. A bad case is the second case of Table~\ref{tab:casest}. ICL-SC didn't make any improvements over the baseline. TCL-SG even got out of control.

Generated summaries in Table~\ref{tab:casens} cover different parts of information in the original document. The vanilla output is just a reordering of the first three sentences.
ICL-SC did better by omitting too detailed content compared to the two baselines. 

In a word, the results show that \textbf{ICL-SC can capture the characteristics of different tasks and do better language modeling}. Besides, by comparing the improvements among these five tasks with different output length, we conclude that our \textbf{ICL-SC is more competitive with tasks having shorter outputs.}
Long outputs, such as summaries in news summarization, bring additional difficulties on the arrangement of multiple salient contents and cross-sentence relations, which can't be well solved with such a simple in-sample curriculum and will be considered in the future.  






\section{Analysis}

For a better understanding of ICL-SC, 
we did comprehensive ablation studies and combined it with the traditional CL. The experiments in this section are done on dialogue summarization, which is representative due to the medium output length.

\subsection{Ablations on the Training Strategy}

To examine the design of decreasing the prefix for ICL-SC, we introduce the alternatives as follows:
\begin{itemize}
	\item \textbf{Decrease} refers to the Algorithm~\ref{alg:picl}. Taking $p_{start}=0.6$ and $s=0.3$ as an example, the prefix percentage $p$ varies as $0.6\rightarrow 0.3\rightarrow 0.0$ during training.
	\item \textbf{Increase} means that we gradually increase the length of prefix by increase $p$ following $0.0\rightarrow0.3\rightarrow0.6$.
	\item \textbf{Random} is that we randomly pick $p$ from the set $\{0.0, 0.3, 0.6\}$ in this example.
\end{itemize}

\begin{table}[h]
	\scriptsize
	\centering
	\begin{tabular}{cccccc}
		\hline
		{Strategy} & {R1} & {R2} & {RL} & {Met} & {BertS} \\
		\hline
		Decrease &\textbf{53.07} & \textbf{28.23} & \textbf{43.83} & \textbf{26.12} & \textbf{72.17}\\
		Increase & 51.43 & 27.35 & 42.97 & 24.32 & 71.25 \\
		Random & 51.80 & 27.69 & 43.27 & 24.59 & 71.51 \\
		\hline
	\end{tabular}
	\caption{Ablations on ICL strategies. The starting point and the stride are 0.6 and 0.3 respectively.}
	\label{tab:ablstrategy}
\end{table}

The results are shown in Table~\ref{tab:ablstrategy}, with Decrease ranking first and Increase ranking the worst.
Decrease significantly outperforms other ablations, showing that our sequence completion criterion of shrinking the prefix does work by means of learning from easy to hard.

\subsection{Parameter Search of the Starting Point and the Stride}
\label{sec:params}

\begin{figure}[h]
	\centering
	\begin{minipage}[t]{0.5\linewidth}
		\centering
		\subfloat[Starting Point]{
			\includegraphics[scale=0.45]{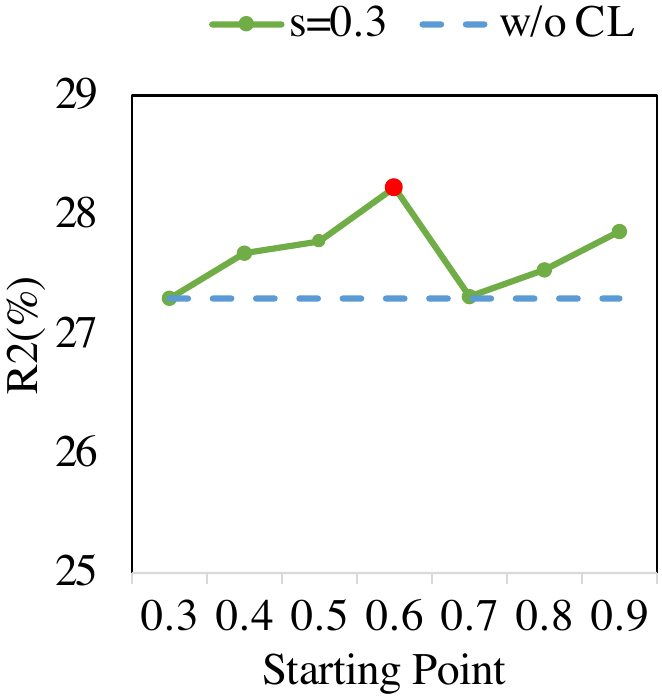}
		}%
	\end{minipage}%
	\begin{minipage}[t]{0.5\linewidth}
		\centering
		\subfloat[Stride]{
			\includegraphics[scale=0.45]{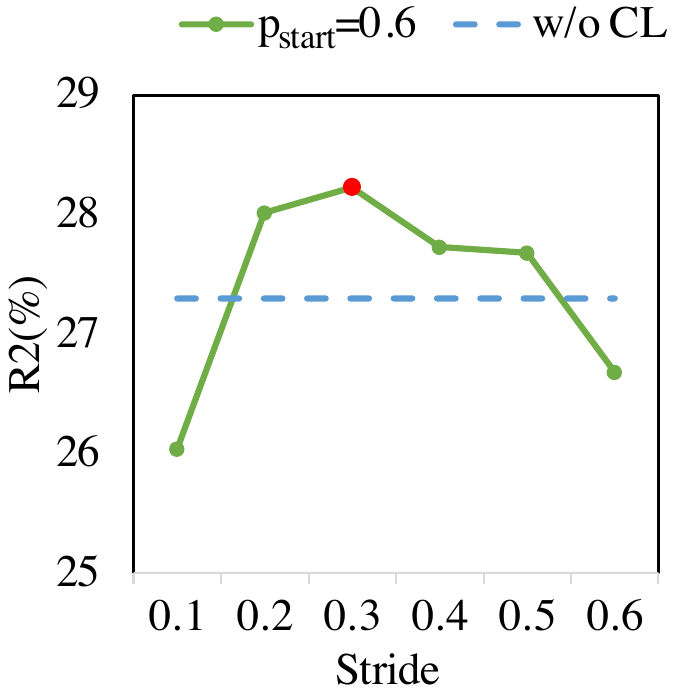}
		}%
	\end{minipage}%
	\centering
	\caption{Parameter search of the starting point $p_{start}$ and the stride $s$. The ``w/o'' CL representing the BART baseline is drawn for comparison.} 
	\label{fig:stridestart}
\end{figure}

To better understand how the ICL-SC manipulates the difficulty of samples during the training process, we further did experiments on different settings of two newly-introduced hyper-parameters $p_{start}$ and $s$. The results are in ~\figref{fig:stridestart}.

We can see that the performance drops with either a too large or too small $p_{start}$. The former one starts training with only predicting the last 1 or 2 tokens according to the average length of reference output shown in Table~\ref{tab:taskdata}. Most of the time, they are punctuation marks that do not carry any important semantic information, leading to a bad warm-up. The latter one requires the model to predict more than half of the output, which are too difficult as a beginning learning target. Besides, a larger $p_{start}$ which is divisible by $s$ is more competitive.

The trend is the same for using different stride values. The performance drops with $s$ equaling 0.1 or 0.6. 
The smaller ones lead to too tiny changes, which not only excessively prolongs the required training time but also leads to server outfitting on the training set. The larger ones greatly enlarge the gap between training targets which degrades to 0.0 directly. It also harms the performances.

In a word, the training should start with a medium difficulty training objective and the gap between training objectives shouldn't be too large. Both parameters are closely related to the output length of different tasks. We suggest using ($p_{start}=0.6$, $s=0.3$) for NLG tasks with multi-sentence outputs, and ($p_{start}=0.5$, $s=0.5$) for NLG tasks with single-sentence outputs. All of our experiments are done based on this guideline. 



\subsection{Combinations with the Traditional CL}

Since our ICL-SC is orthogonal to sample-wise CL and designing an appropriate sample-wise curriculum is not easy, we choose dialogue summarization as a representative task, design several traditional CL strategies empirically, and further apply our ICL-SC on top of them for comparisons.
$4$ different traditional CL strategies are as follows:
\begin{itemize}
	\item \textbf{Input length (InLen)} refers to the number of tokens in the input dialogue. The longer a dialogue is, the more complex a sample is.
	\item \textbf{Output length (OutLen)} is the number of tokens in a reference summary, which is also proportional to the difficulty of a sample.
	\item \textbf{Compression ratio (CompR)} equals the output length divided by the input length. More compressed training pairs are harder.
	\item \textbf{Abstractiveness (Abstr)} represents the percentage of novel words in the reference summary which are not in the dialogue. We measure it by Rouge-2 recall, which is inversely proportional to the difficulty level.
\end{itemize}

\begin{table}[th]
	\scriptsize
	\centering
	\begin{tabular}{lccccc}
		\hline
		{Method} & {R1} & {R2} & {RL} & {Met} & {BertS} \\
		\hline
		w/o CL & 51.88 & 27.30 & 42.77 & 24.75 & 71.38 \\
		ICL-SC & {\textbf{53.07}} & \textbf{28.23} & \textbf{43.83} & {\textbf{26.12}}& {\textbf{72.17}} \\
		\hline
		InLen & 52.19 & \textbf{27.73} & \textbf{43.50} & 25.57 & 71.73\\
		InLen+ & \textbf{52.56} & 27.60 & 43.43 & \textbf{25.77} & \textbf{71.92}\\
		\hline
		OutLen & 41.38 & 20.88 & 31.77 & \textbf{27.95} & 67.21\\
		OutLen+ &\textbf{43.96} & \textbf{22.14} & \textbf{33.05} & 26.39 & \textbf{67.64} \\
		\hline
		CompR & 39.68 & 19.28 & 34.73 & 14.41 & 65.96 \\
		CompR+ & \textbf{41.59} & \textbf{20.78} & \textbf{36.62} & \textbf{15.22} & \textbf{67.19}\\
		\hline
		Abstr & \textbf{44.61} & 20.10 & 36.93 & \textbf{17.34} & 68.29 \\
		Abstr+ & 44.41 & \textbf{20.64} & \textbf{37.29} & 17.25 & \textbf{68.33} \\
		\hline
	\end{tabular}
	\caption{Performaces with traditional CL strategies. ``+'' represents experiments further armed with ICL-SC.}
	\label{tab:traditional}
\end{table}
The results based on the ordered training samples according to these intuitive CL strategies are shown in Table~\ref{tab:traditional}. It shows that only InLen improves the vanilla model, but it still lags behind the pure ICL-SC. Other strategies failed mainly due to the low data quality at the beginning or the end of training. 
Taking Abstr as an example, samples with the highest 
Rouge-2 recall are gathered at the beginning where 
their inputs and outputs are almost the same. 
This leads to a bad initialization for models learning 
the summarization ability. 

Besides, some strategies 
are incompatible, such as OutLen and CompR. Samples with the shortest output length are always too compressed. Therefore, developing a comprehensive score for a better ranking is difficult. It should be also noticed that most of these strategies are designed for summarization, which are not suitable for generalization.

In a word, it's hard to develop a 
comprehensive strategy for one task or a unified strategy for different NLG tasks with traditional CL. 
ICL-SC not only outperforms these CL strategies, but also improves them when easily combined. 

\label{sec:tracl}

\section{Related Work}
\textbf{Natural language generation} has received great attention with deep neural networks, especially pre-trained language models. It refers to the task where expected outputs for different purposes are in natural language~\cite{dong2021survey}. The inherent characteristic of having more than one correct output given the same input is the core challenge of solving this kind of task, especially for evaluation~\cite{singh2018does}.

\textbf{Curriculum learning}~\cite{bengio2009curriculum} boost models' performances in a range of machine learning areas~\cite{LiuGW20,varshney2022model} by reordering the training samples.
It meets great obstacles when applying to NLG tasks as it's hard to evaluate the difficulties of training samples. 
Different rules are developed for different tasks~\cite{platanios2019competence,chang2021does}. For example, \cite{liu2018curriculum} measures the complexity of question-answering pairs from the view of frequency and grammar simply for answers. \cite{kocmi2017curriculum} focuses more on POS features and the length of translation pairs.
Other works utilize additional models or targeting models in the previous training step~\cite{zhang-etal-2019-curriculum,zhang2018empirical}. \citet{shen2020cdl} reorder samples by the accuracy from an independent emotion classifier for response generation. However, such salient features do not always exist or can be well classified. There is also work~\cite{zhou2020uncertainty} 
using either the reference perplexity or generations evaluated by corresponding metrics for ranking during training, while these scores are not ideal due to the one-to-many characteristic of NLG. 
Thus, designing a CL strategy generalizing well for NLG is difficult.

Instead of figuring out the oracle scoring function for training samples, 
we propose to measure the language generation difficulty within a sample. 
\citet{liang-etal-2021-token-wise} did something similar though their approach
amounts to data augmentation by doing sub-sequence generation, which is
not exactly curriculum learning. We, on the other hand,  train on the original sample with a decreasing prefix length and thus learn from easy to hard.

\section{Conclusion}

This paper defines a kind of curriculum learning strategy for NLG tasks called in-sample curriculum learning (ICL) by manipulating the difficulty of training within a training sample instead of ranking among samples. We propose the ICL algorithm with the sequence completion curriculum which boosts the performance of strong baselines on a wide range of tasks, showing the effectiveness and strong generalization ability of our approach. More training strategies under ICL digging the inherent difficulties of generating a language sequence are expected in the future.


\section*{Limitations}

\begin{table} [h]
	\scriptsize
	\centering
	\begin{tabular}{p{2.5cm}p{1cm}p{1cm}p{1cm}}
		\hline
		{Tasks} & {w/o CL} & {TCL-SG} & {ICL-SC} \\
		\hline
		Reading Comprehension  &6.67 ep & 13.00 ep & 7.67 ep \\
		Dialog Summarization &6.00 ep &15.67 ep & 11.67 ep\\
		Style Transfer &6.50k st & 14.78k st & 9.67k st  \\
		Question Generation & 17.67k st& 37.73k st  & 21.00k st  \\
		News Summarization &21.00k st  & 47.20k st &36.00k st\\
		
		\hline
	\end{tabular}
	\caption{Average number of training steps for different approaches. ``ep'' and ``st'' are short for ``epochs'' and ``steps'' respectively.}
	\label{tab:limitations}
\end{table}

One limitation of our approach is that in-sample curriculum learning methods (both TCL-SG and ICL-SC) always incur extra overhead during training compared with the vanilla model shown in Table~\ref{tab:limitations}. Nevertheless, the inference time of different approaches is the same as the vanilla model.
In a word, it's worthwhile because (1) ICL-SC can perform significantly better than both baselines without additional computational requirements during inference in real applications; (2) ICL-SC doesn't rely on task-specific expertise and has strong generalization ability.

Due to the limited computational resources, we were unable to do experiments on machine translation.
According to the implementation details in \citet{liang-etal-2021-token-wise}, all of their machine translation experiments were done on 32G NVIDIA V100 GPUs which are much more powerful than a single RTX 3090.
Even for the low resource setting with around 133K to 612K training samples, they used dynamic batching with 4096 maximum tokens and trained for 60 epochs.
This will either lead to an out-of-memory error or take us several weeks or even months to get the results of a single run on our machine.
Instead, we tried our best to cover a range of representative natural language generation tasks and corresponding datasets with different characteristics, such as sizes and output lengths (\tabref{tab:taskdata}).

\section*{Acknowledgments}
This work was generously supported by the CMB Credit Card Center \& SJTU
joint research grant, and Meituan-SJTU joint research grant.

\bibliography{acl23}
\bibliographystyle{acl_natbib}


\appendix

\section{Packages used for Baselines}
\label{sec:pkgs}

The packages we adopted to re-implement the baseline are listed as follows:

\textbf{Reading Comprehension}
\begin{itemize}
	\item Dataset: \url{https://github.com/nlpdata/dream/tree/master/data}
	\item Baseline Code: \url{https://github.com/huggingface/transformers}
	\item Evaluation Metric: \url{https://github.com/tensorflow/nmt/blob/master/nmt/scripts/bleu.py}
\end{itemize}

\textbf{Dialogue Summarization}
\begin{itemize}
	\item Dataset: \url{https://arxiv.org/src/1911.12237v2/anc/corpus.7z}
	\item Baseline Code: \url{https://github.com/huggingface/transformers}
	\item Evaluation Metric: \url{https://github.com/pltrdy/files2rouge}; \url{https://github.com/Yale-LILY/SummEval}
\end{itemize}

\textbf{Style Transfer}
\begin{itemize}
	\item Dataset: \url{https://github.com/martiansideofthemoon/style-transfer-paraphrase}
	\item Baseline Code: \url{https://github.com/martiansideofthemoon/style-transfer-paraphrase}
	\item Evaluation Metric: \url{https://github.com/martiansideofthemoon/style-transfer-paraphrase}
\end{itemize}

\textbf{Question Generation}
\begin{itemize}
	\item Dataset: \url{https://github.com/microsoft/unilm/tree/master/unilm-v1}
	\item Baseline Code: \url{https://github.com/microsoft/unilm/tree/master/unilm-v1}
	\item Evaluation Metric: \url{https://github.com/microsoft/unilm/tree/master/unilm-v1}
\end{itemize}

\textbf{News Summarization}
\begin{itemize}
	\item Dataset: \url{https://drive.google.com/file/d/0BzQ6rtO2VN95a0c3TlZCWkl3aU0/view?resourcekey=0-toctC3TNM1vffPCZ7XT0JA}
	\item Baseline Code: \url{https://github.com/huggingface/transformers}
	\item Evaluation Metric: \url{https://github.com/pltrdy/files2rouge}; \url{https://github.com/Yale-LILY/SummEval}
\end{itemize}

\section{Preliminary Studies on TCL}
\label{sec:preliminary}

Preliminary studies on dialogue summarization for TCL under different settings are shown in Table~\ref{tab:tclpre}.
We can see that the ``soft'' setting does help the TCL with sub-sequence generation curricula, which is consistent with the results in \citet{liang-etal-2021-token-wise}.
Results are opposite for TCL with our proposed sequence completion curricula. The ``soft'' setting considering the loss from prefix tokens actually hurts the intuition that ``the shorter the target is, the easier the tasks is''.  As a result, SC-hard performs better than SC-soft.

\begin{table}[th]
	\scriptsize
	\centering
	\begin{tabular}{cccccc}
		\hline
		{} & {R1} & {R2} & {RL} & {Met} & {BertS} \\
		\hline
		w/o CL& 51.88 & 27.30 & 42.77 & 24.75 & 71.38 \\
		\hline
		SG-hard &50.70 & 27.31 & 43.00 & 23.47 & 70.85\\
		SG-soft & 52.43 & 27.65 & 43.56 & 25.17 & 71.86 \\
		\hline
		SC-hard &  52.69 & 28.28 & 43.89 & 25.08 & 71.95 \\
		SC-soft & 51.39 & 27.53 & 43.06 & 23.84 & 71.35 \\
		\hline
	\end{tabular}
	\caption{Ablations on TCL learning algorithm with different settings.}
	\label{tab:tclpre}
\end{table} 

Experiments on the sensitivity of curriculum step in TCL-SG~\cite{liang-etal-2021-token-wise} are in Table~\ref{tab:tclpre2}. 
It consistently has improvements on dialogue summarization compared with the baseline. However, the performances also vary a lot with different curriculum steps, especially on R1, Meteor and BertScore. The estimation rule proposed in~\citet{liang-etal-2021-token-wise} of computing the number of steps it takes to reach approximately 70\% of final scores doesn't perform well for dialogue summarization. So, we choose to set curriculum steps to 3 epochs for dialogue summarization and news summarization, and 2 epochs for reading comprehension and style transfer, which not only achieve better results, but also are fairer for comparisons. For news summarization, we still adopted their estimation rule and trained with 5200 curriculum steps.

\begin{table}[th]
	\scriptsize
	\centering
	\begin{tabular}{cccccc}
		\hline
		{Curriculum Step} & {R1} & {R2} & {RL} & {Met} & {BertS} \\
		\hline
		w/o CL & 51.88 & 27.30 & 42.77 & 24.75 & 71.38 \\
		\hline
		1 epoch & 52.48 & 27.86 & 43.47 & 25.50 &71.83 \\
		1.58 epoch(70\%) & 51.89 & 27.64 & 43.51 & 24.37 &71.55 \\
		2 epoch & 51.93 & 27.75 & 43.37 & 24.73 & 71.57\\
		3 epoch & 52.43 & 27.65 & 43.56 & 25.17 & 71.85\\
		\hline
	\end{tabular}
	\caption{Performances on TCL-SG with different curriculum steps.}
	\label{tab:tclpre2}
\end{table}







\end{document}